\documentclass[letterpaper, 10 pt, conference]{ieeeconf}  

\IEEEoverridecommandlockouts                              
\usepackage[utf8]{inputenc}
\usepackage[T1]{fontenc}

\usepackage{graphics} 
\usepackage{amsmath} 
\usepackage{multirow}
\usepackage{booktabs}
\usepackage{graphicx}
\usepackage{caption}
\usepackage{pifont}

\usepackage{colortbl}

\usepackage[table,xcdraw]{xcolor}
\usepackage{amssymb}

\definecolor{rblue}{rgb}{0,0.5,1}
\definecolor{awesome}{rgb}{1.0, 0.13, 0.32}
\definecolor{hollywoodcerise}{rgb}{0.96, 0.0, 0.63}
\definecolor{lasallegreen}{rgb}{0.03, 0.47, 0.19}
\definecolor{hanpurple}{rgb}{0.32, 0.09, 0.98}
\definecolor{green(pigment)}{rgb}{0.0, 0.65, 0.31}

\definecolor{mygray}{gray}{.9}
\definecolor{mygreen}{RGB}{93,174,86}

\makeatletter
\newcommand{\thickhline}{%
	\noalign {\ifnum 0=`}\fi \hrule height 1pt
	\futurelet \reserved@a \@xhline
}
\makeatother

\makeatletter
\let\NAT@parse\undefined
\makeatother
\usepackage[pagebackref=false,breaklinks=true,colorlinks,bookmarks=false]{hyperref}
\hypersetup{colorlinks=true,linkcolor={red},citecolor={hanpurple},urlcolor={magenta}}

\title{\LARGE \bf
Resource-Efficient Affordance Grounding with Complementary Depth and Semantic Prompts
}

\author{Yizhou Huang$^{1}$, Fan Yang$^{1}$, Guoliang Zhu$^{1}$, Gen Li$^{2}$, Hao Shi$^{3}$, Yukun Zuo$^{1}$,\\Wenrui Chen$^{1}$, Zhiyong Li$^{1,*}$, and Kailun Yang$^{1,*}$
\thanks{This work was supported in part by the National Key R\&D Program (Grant No. 2022YFB4701400/2022YFB4701404), the National Natural Science Foundation of China (Grant No. 62473139), in part by the Hunan Provincial Research and Development Project (Grant No. 2025QK3019), and in part by the Open Research Project of the State Key Laboratory of Industrial Control Technology, China (Grant No. ICT2025B20).}
\thanks{$^{1}$The authors are with the School of Artificial Intelligence and Robotics and the National Engineering Research Center of Robot Visual Perception and Control Technology, Hunan University, China (email: zhiyong.li@hnu.edu.cn, kailun.yang@hnu.edu.cn).}%
\thanks{$^{2}$The author is with the School of Informatics, University of Edinburgh, UK.}%
\thanks{$^{3}$The author is with the State Key Laboratory of Extreme Photonics and Instrumentation, Zhejiang University, China.}%
\thanks{*Corresponding authors: Zhiyong Li and Kailun Yang.}%
}

\begin{document}

\maketitle
\thispagestyle{empty}
\pagestyle{empty}

\begin{abstract}
Affordance refers to the functional properties that an agent perceives and utilizes from its environment, and is key perceptual information required for robots to perform actions. This information is rich and multimodal in nature. Existing multimodal affordance methods face limitations in extracting useful information, mainly due to simple structural designs, basic fusion methods, and large model parameters, making it difficult to meet the performance requirements for practical deployment. To address these issues, this paper proposes the BiT-Align image-depth-text affordance mapping framework. The framework includes a Bypass Prompt Module (BPM) and a Text Feature Guidance (TFG) attention selection mechanism. BPM integrates the auxiliary modality depth image directly as a prompt to the primary modality RGB image, embedding it into the primary modality encoder without introducing additional encoders. This reduces the model's parameter count and effectively improves functional region localization accuracy. The TFG mechanism guides the selection and enhancement of attention heads in the image encoder using textual features, improving the understanding of affordance characteristics. Experimental results demonstrate that the proposed method achieves significant performance improvements on public AGD20K and HICO-IIF datasets. On the AGD20K dataset, compared with the current state-of-the-art method, we achieve a $6.0\%$ improvement in the KLD metric, while reducing model parameters by $88.8\%$, demonstrating practical application values. The source code will be made publicly available at~\url{https://github.com/DAWDSE/BiT-Align}.
\end{abstract}


\section{Introduction}

``Affordances'' refer to the functional properties that an agent perceives and utilizes from its environment~\cite{gibson1977theory}. This concept is widely used in cognitive science to understand and explain the correspondence between object properties and agent behaviors in the environment. As a crucial component of embodied intelligence, affordances serve as an important bridge that connects object perception and agent actions. In the field of computer vision, accurately identifying the regions where an agent interacts with objects, such as blades used for cutting or handles used for grasping, has been a major focus of research, with significant applications in robot grasping~\cite{puang2024learningstablerobotgrasping} and scene understanding~\cite{delitzas2024scenefun3d}.

\begin{figure}[!t]
    \centering
    \includegraphics[width=0.48\textwidth]{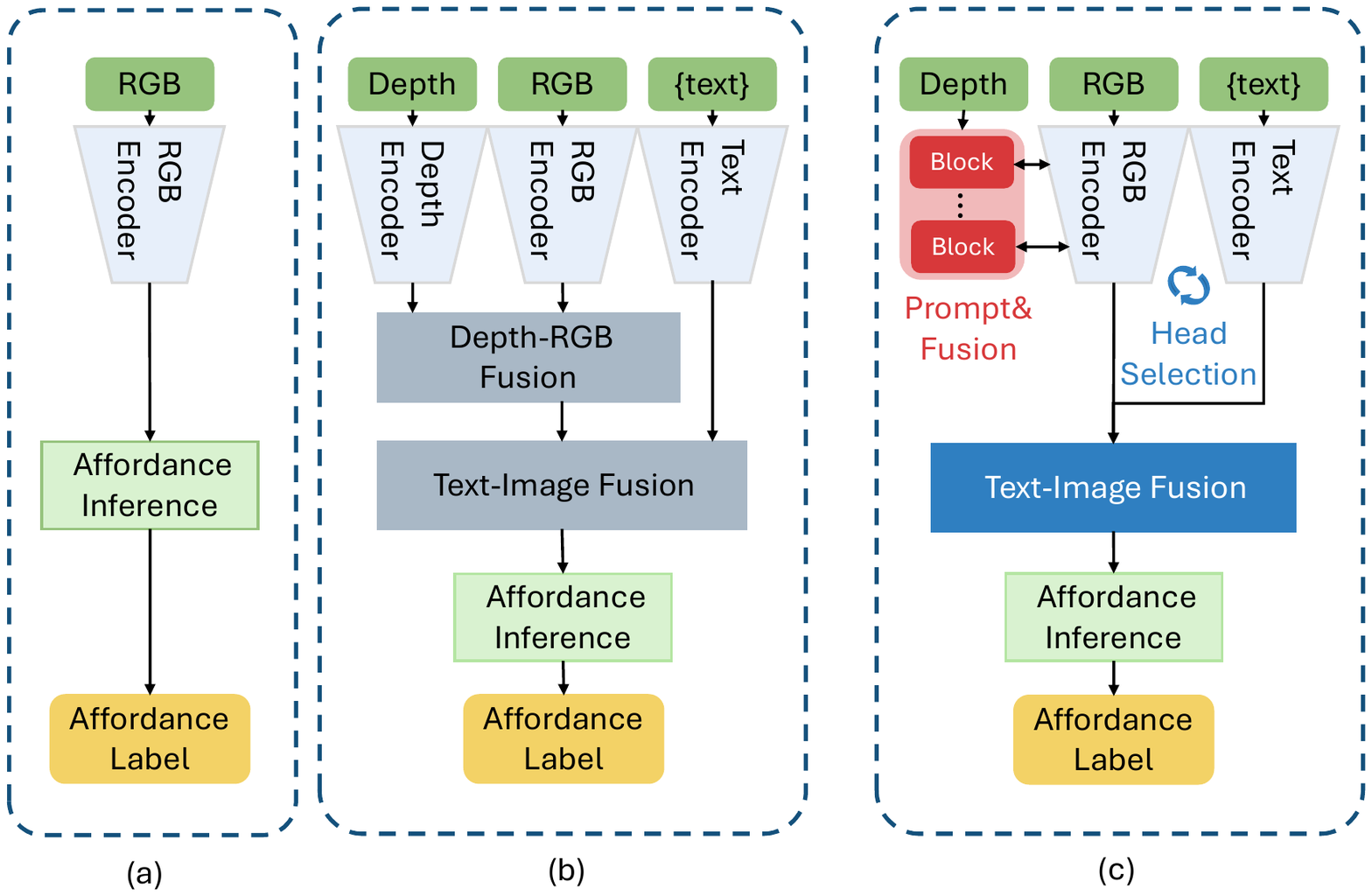} 
    \captionsetup{font=small}
    \vskip-2ex
    \caption{\textbf{Comparison of typical methods (a) (b) and the framework proposed in this paper (c).} Compared to unimodal method (a), as well as multimodal method (b) that employs multiple encoders and then performs fusion, our method (c) simultaneously achieves feature encoding and fusion through a bypass module, and also proposes an image-text fusion optimization method utilizing attention heads. }
    \label{fig:wide-figure} 
    \vskip-3ex
\end{figure}

Early affordance research predominantly relied on vision-based neural network methods, utilizing RGB unimodal features~\cite{chuang2018learning, nagarajan2019grounded, luo2023learning, geng2023gapartnet} to extract affordance-related information from images or videos, as shown in Fig.~\ref{fig:wide-figure}(a). These methods attempt to capture visual features related to object functionality by analyzing the geometric texture of objects. However, relying solely on RGB geometric texture information is insufficient to comprehensively understand the ``actionable regions'' of an object, \textit{i.e.}, its affordances. In fact, human ``language instructions'' and ``structural information'' of objects also contain important affordance cues. For example, when we see a child grasping the handle of a tennis racket to perform a ``hit'', and then see a badminton racket or ping-pong paddle, we instinctively guide the child to grip the handle of these objects and perform a similar ``hit'' action. This language-based guidance is part of the affordance cues. Additionally, when we know that the switch and handle of a drill require the index finger and the other four fingers to work together, we can also recall the corresponding actions when seeing a watering can with similar switch and handle structural features. This reflects the affordance cues embedded in the ``structure'', and such structural information is often reflected in the depth information of the object.

In recent years, several studies have begun to explore multimodal affordance research by integrating images, depth, and textual information. For example, the work in~\cite{gao2024learning2d} proposed a multi-image guided 3D affordance grounding framework, which identifies common interaction patterns across multiple human-object interaction images to ground 3D object affordance regions with point cloud and RGB images. 
Furthermore, the work in~\cite{tong2024ovalprompt} developed a prompt-based approach for open-vocabulary affordance localization in RGB-D images. However, existing multimodal fusion methods typically encode each modality separately and then fuse them, as shown in Fig.~\ref{fig:wide-figure}(b). While intuitive, this fusion process is rather simplistic, extracting only limited useful features while simultaneously increasing the model's parameter count, which leads to significant computational overhead and efficiency issues when deployed in real-world agent systems.

\begin{figure}[!t]
  \centering
  \includegraphics[width=0.48\textwidth]{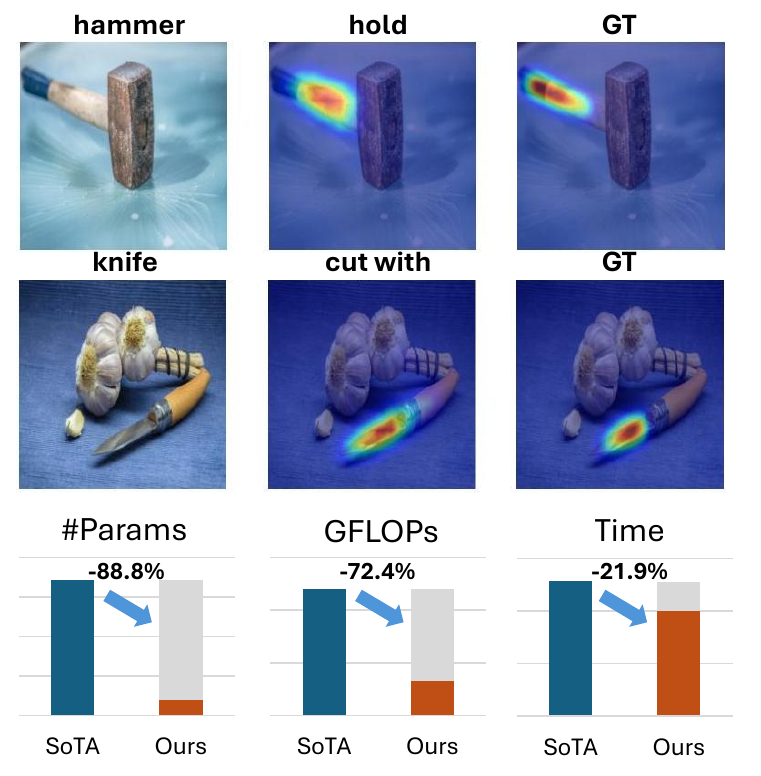}
  \vskip-2ex
  \caption{Visualization results of our method and efficiency improvement of our proposed method compared to the State-of-The-Art (SoTA) affordance grounding method~\cite{xu2024weakly}.
  }
  \label{fig:teaser}
  \vskip -3ex
\end{figure}

To address these issues, this paper proposes the method BiT-Align: Bypass and Text-guided Attention for Multimodal Affordance Grounding. As shown in Fig.~\ref{fig:wide-figure}(c), the proposed Bypass Prompt Module (BPM) and Text Feature Guidance (TFG) attention selection mechanism are effectively deployed to address the problem of aligning depth information, visual part features, and textual features. 

Specifically, we first design BPM, which directly integrates the auxiliary-modality depth image as a prompt to the primary-modality RGB image encoder, avoiding the need for an additional encoder. 
This integration reduces computational burden while improving the accuracy of functional region localization by incorporating depth information into the RGB image encoding process. 
Secondly, we introduce the TFG attention selection mechanism, which matches the attention head masks from the intermediate layers of the image encoder with the textual features, thereby guiding textual information into the process of selecting and enhancing image features. 
During the output phase, we use the weighted attention head masks to highlight and enhance the corresponding image regions that are related to the textual input. 
Finally, we generate the final fused features using a text-image fusion module. 
These fused features are then input into the inference network to achieve accurate and efficient affordance localization.

The proposed BiT-Align model is verified via a comprehensive variety of experiments and ablation studies on public AGD20K~\cite{luo2022learning} and HICO-IIF~\cite{xu2024weakly} datasets.
While achieving state-of-the-art affordance grounding performance as shown in Fig.~\ref{fig:teaser}, the model's parameter amount is reduced by $88.8\%$ compared to the current multimodal method~\cite{xu2024weakly}, with a $6.0\%$ reduction in the KLD metric (lower is better). Compared to the unimodal baseline~\cite{li2023locate}, the proposed model requires only $19.2\%$ additional parameters, making it highly efficient for practical deployment.

The contributions of this work are as follows:
\begin{itemize}
    \item A BiT-Align image-depth-text affordance mapping framework is proposed, which combines RGB, depth, and textual information to achieve efficient multimodal affordance learning.
    \item Bypass Prompt Module (BPM) is proposed to integrate the auxiliary modality depth image as a prompt to the primary modality RGB image encoder, reducing model parameters and improving functional region localization accuracy without the need for additional encoders.
    \item A Text Feature Guidance (TFG) attention selection mechanism is proposed, which enhances the understanding of affordance features by guiding the selection and enhancement of attention heads in the image encoder using textual features.
    \item Extensive experiments on multiple affordance datasets demonstrate that the proposed method significantly outperforms existing approaches in affordance localization and segmentation with efficiency enhancement.
\end{itemize}

\section{Related Work}
In this work, we propose a deep feature prompting and text attention-based affordances grounding method. 
In this section, we will summarize previous work related to and inspiring our approach.

\subsection{Visual Affordance Grounding}
The concept of affordance grounding originates from biology, aiming to identify interactive regions of objects and their associated functionalities. Early works~\cite{koppula2013learning, kokic2017affordance, myers2015affordance} treated affordance grounding as a pixel-level segmentation task and employed fully supervised training. 

To address the challenges of annotation, many researchers began to draw inspiration from human learning mechanisms and proposed weakly supervised training using interaction videos or photos. Following this, the work in~\cite{hou2021affordance} focused on transferring affordance knowledge from third-person to first-person perspectives. With the emergence of self-supervised feature extraction models~\cite{caron2021emerging}, the study~\cite{luo2022learning} established a training framework for `egocentric' and `exocentric' views and proposed a knowledge transfer mechanism between the two training branches. To further optimize training efficiency and reduce parameters, LOCATE~\cite{li2023locate} employed clustering to separate foreground and background in views and extract affordance regions, whereas Luo~\textit{et al.}~\cite{luo2023leverage} mapped object affordances to human body parts using keypoints. 
In addition, a piece of recent research in~\cite{ragusa2023affordance} explored the use of compact networks for wearable devices, and recent works of~\cite{yang2023recent,chu2019learning,chu2019toward} conducted experiments in robotic arms, highlighting the challenges faced in applying affordance tasks to robotic arms. 
Current work has established an effective and parameter-efficient affordance training framework, providing highly favorable conditions for further improvements based on this foundation.
On this basis, we introduce deep modal information to improve the understanding of the affordance of the model in the spatial shape.

\subsection{Multimodal Affordance Grounding}
To enrich training information and enhance the performance of affordance grounding, many studies have begun to integrate textual information into affordance localization. 
The method in~\cite{chen2024worldafford} optimized the LOCATE method by using SAM masks and matching them with textual features, while also utilizing LLM chain-of-thought prompting to assist affordance learning. Similarly, the method in~\cite{yoshida2024text} used LLM models to obtain textual prompts to aid in affordance learning, and INTRA~\cite{jang2025intra} introduced interaction relationship-guided contrastive learning with LLM, enabling the model to perform without the need for an egocentric training branch. 
Meanwhile, the work in~\cite{li2024one} achieved outstanding one-shot open-vocabulary performance by combining CLIP~\cite{radford2021learning} and DINO-V2~\cite{oquab2023dinov2} encoded features. 
Additionally, the work in~\cite{xu2024weakly} provided a novel perspective by using attention weights to complement textual features. 
To address the scarcity of multimodal training datasets, some studies, such as AffordanceLLM~\cite{qian2024affordancellm} and GAT~\cite{li2024learning}, leveraged depth estimation~\cite{yang2024depth} to acquire additional modalities for assistance. 
The introduction of multimodal information improves model accuracy but inevitably increases the number of parameters, raising the costs of model training and inference. 

\begin{figure*}[!t]
  \centering
  \includegraphics[width=0.8\textwidth]{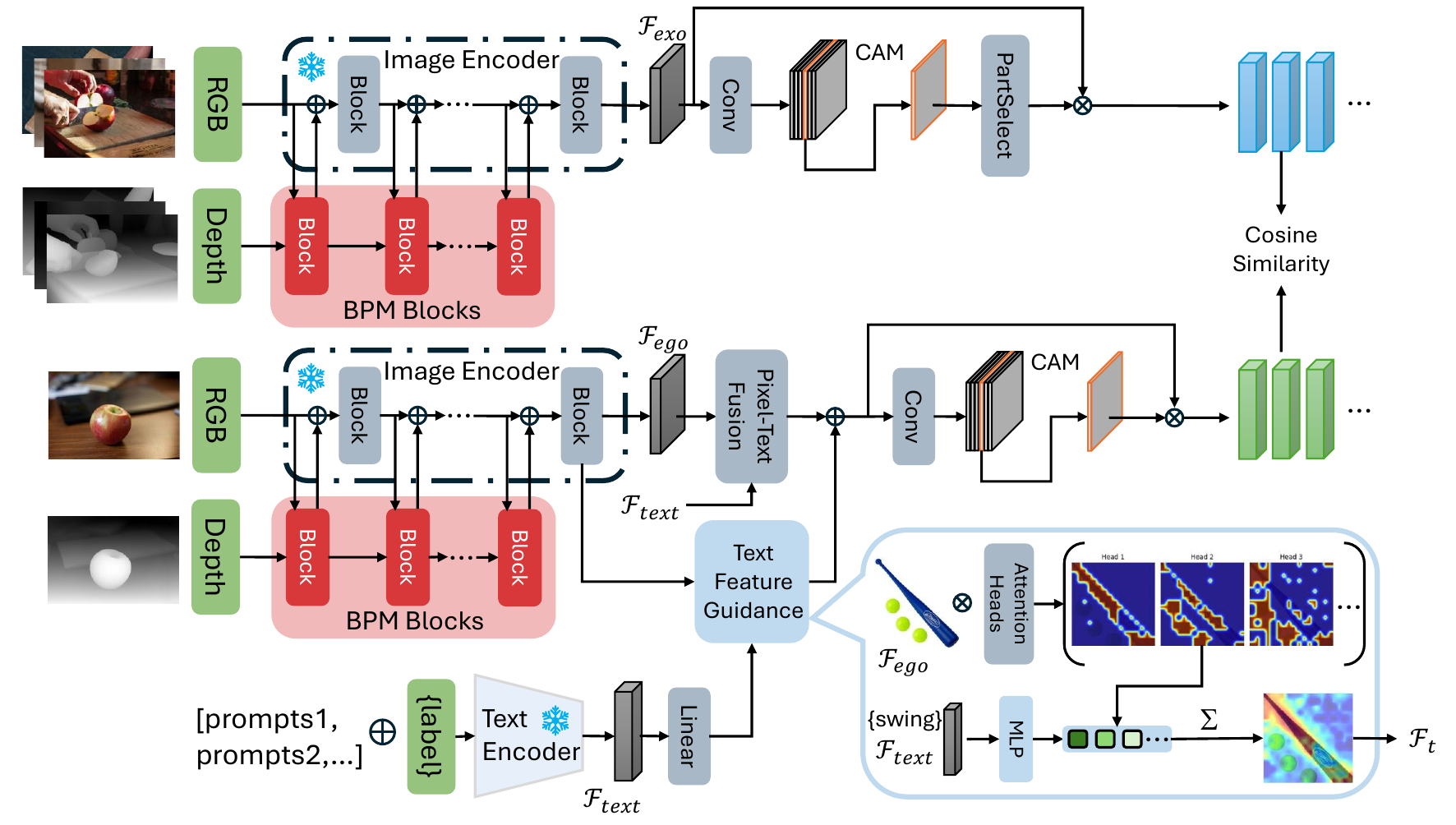}
  \vskip-1ex
  \caption{\textbf{Overview of the proposed BiT-Align framework for image-depth-text affordance mapping.} 
  The egocentric branch (ego), exocentric branch (exo), and text branch, form the overall framework of the model. 
  Only the egocentric branch and text branch are retained in the inference stage.
  The red blocks form the Bypass Prompt Module (BPM), receiving input from depth images and progressively fusing features into the RGB encoder. 
  The designed Text Feature Guidance (TFG) and the Pixel-Text (PT) fusion modules~\cite{xu2024weakly} both receive the same text features and RGB-D fused features, and then perform fusion in respective methods.
  In the framework, the weights of the image and text encoders are frozen.
  }
  \label{fig:framework}
  \vskip-3ex
\end{figure*}

\subsection{Prompt Learning}
The role of prompting in the field of Natural Language Processing (NLP) has been extensively studied and has been proven effective. 
As NLP methods are increasingly introduced into the vision domain, prompting techniques have gradually gained significant attention from researchers~\cite{xin2024parameter}. 
The research in~\cite{jie2024convolutional} aimed to leverage prompting to integrate the contextual capabilities of transformers with the prior knowledge of local features in Convolutional Neural Networks ~(CNNs). 
The study presented in~\cite{chen2022adaptformer} employed a typical gourd-shaped adapter for fine-tuning visual transformers, whereas the approach in~\cite{nie2023pro} further combined CNN-like adapter-based prompting methods to generate more compact and robust models for downstream tasks. 
The work of~\cite{xu2024lv} proposed a parameter-efficient adapter method, and the research in~\cite{dong2024efficient} treated multimodal fusion as a form of feature prompting, eliminating the need for multiple encoders, which facilitates faster model inference and training. 

As a fine-tuning method, prompt learning is characterized by its low parameter count. This opens up possibilities for introducing new modalities into models. Specifically, by using bypasses or adapters, new modalities can be treated as prompts or corrections, supplementing the single modality. 
This approach has the potential to achieve a lower parameter introduction at the cost of some accuracy. We refer to and introduce the method of prompt learning into affordance grounding, by treating one input modality as a prompt for the other input modality.

\section{Method}

\subsection{Overview}

In this paper, we focus on the task of weakly-supervised affordance grounding, where the goal is to extract affordance knowledge from several exocentric human-object interaction images and transfer it to the egocentric counterpart.
During the training process, only the action labels of the exocentric images can be used as weak supervision.
To tackle this task, we propose a resource-efficient framework, BiT-Align, which integrates RGB, depth, and textual information to achieve efficient multimodal affordance grounding.
An overview of the established framework is shown in Fig.~\ref{fig:framework}. 
Specifically, BiT-Align includes two innovative modules: Bypass Prompt Module (BPM) and Text Feature Guidance (TFG).
BPM is designed to incorporate geometric information from depth images, which are important cues for affordance understanding. 
TFG, on the other hand, aims to provide guidance to the image features by leveraging textual information, leading to more discriminative features for the affordance mapping.
The details of these two modules are described in Sec.~\ref{sec:prompt_fusion} and Sec.~\ref{sec:text_feature_guidance}, respectively.

\subsection{Bypass Prompt Module} 
\label{sec:prompt_fusion}

\begin{figure}[!t]
    \centering
    \includegraphics[width=0.48\textwidth]{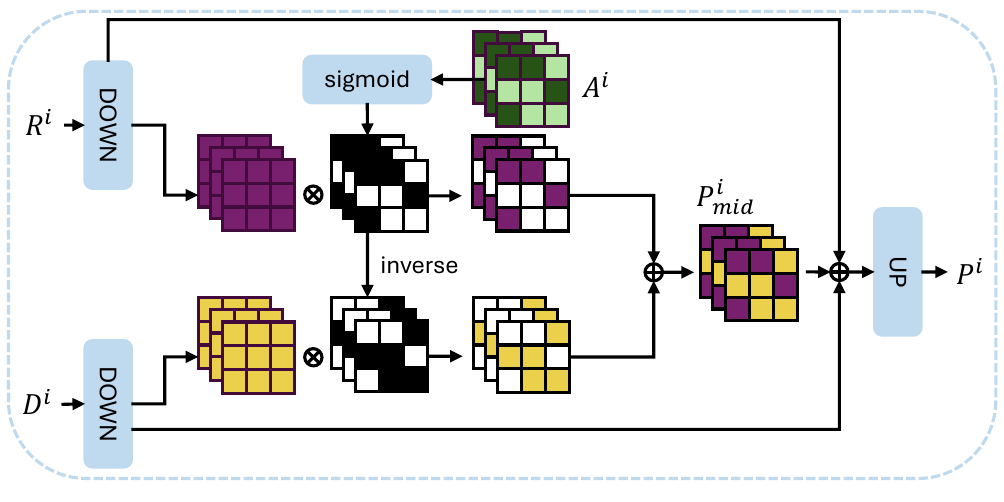} 
    \captionsetup{font=small}
    \vskip-1ex
    \caption{\textbf{Details of the designed BPM block.} The Bypass Prompt Module (BPM) consists of multiple blocks connected in series. Both \textbf{UP} and \textbf{DOWN} are linear layers.}
    \label{fig:fp_fusion} 
    \vskip-3ex
\end{figure}

Unlike prior multimodal fusion methods that employ separate encoders for depth and RGB inputs, the proposed Bypass Prompt Module (BPM) integrates depth information into the intermediate layers of the RGB encoder in a progressive manner.
This design effectively reduces the number of parameters and computational overhead while maintaining the accuracy of functional region localization.
Given the paired exocentric and egocentric images $\mathcal{R}$, we first input them into the vision transformer encoder $\Psi$ to obtain the feature maps $\mathcal{F}_{exo}$ and $\mathcal{F}_{ego}$, respectively:
\begin{equation}
\mathcal{F}_{ego,exo} = \Psi(\mathcal{R}_{ego,exo}).
\end{equation}

Subsequently, during the encoding process, we employ the BPM to fuse the geometric features from the corresponding depth image $\mathcal{D}$ into the intermediate RGB features, and add the result back to the main modality branch, as shown in Fig.~\ref{fig:fp_fusion}.
This process is repeated across the blocks in the encoder, with each BPM receiving RGB and depth features from the previous module.
Specifically, the features are first reduced $\beta$ times in dimensionality through two linear layers to obtain $\mathcal{R}_{m}^{i}$ and $\mathcal{D}_{m}^{i}$ ($i{=}0,1,2,...,N$), where $\beta$ represents the downscaling factor, reducing the dimensionality of the feature vector to ${1}/\beta$ of its original size.
These are then multiplied by a sigmoid-binarized matrix $\mathcal{A}^{i}\in\mathbb{R}^{{c \times hw} \times \frac{d}{\beta}}$ or its inverse $\mathcal{\neg{A}}^{i}$. 
$\mathcal{A}^{i}$ is initially a randomly generated weight matrix, which is updated during the training process to adapt to different modalities. Next, the results are summed to $\mathcal{P}_{mid}^{i}$, which is then added to $\mathcal{R}_{m}^{i}$ and $\mathcal{D}_{m}^{i}$.
Finally, the features are projected back to the original dimensions through a linear layer, resulting in the fused feature $\mathcal{P}^{i}$. 
$\mathcal{P}^{i}$ is added back to the main modality branch and also passed to the next BPM block as the depth feature input for subsequent processing:
\begin{equation}
\mathcal{R}_{m}^{i}, \mathcal{D}_{m}^{i} = 
\begin{cases}
{Linear}(\mathcal{R}, \mathcal{D}), & \text{if } i = 0, \\
{Linear}(\mathcal{R}_{m}^{i-1}, \mathcal{D}_{m}^{i-1}). & \text{if } i > 0.
\end{cases}
\end{equation}
\begin{equation}
\mathcal{P}_{mid}^{i} = \sigma\mathcal{A}^{i}\cdot\mathcal{R}_{m}^{i} + \sigma{\neg\mathcal{A}}^{i}\cdot\mathcal{D}_{m}^{i},
\end{equation}
\begin{equation}
\mathcal{P}^{i} = {Linear}(\mathcal{P}_{mid}^{i} + \mathcal{R}_{m}^{i} + \mathcal{D}_{m}^{i}).
\end{equation}

In this processing step, $\mathcal{A}$ acts as a filter to control the integration of depth information into the RGB features. This allows the model to emphasize depth characteristics at relevant locations, obtaining sufficiently fine-grained information, while still relying on RGB features for decision-making at other positions. 
Besides the features themselves, intermediate quantities from the computation are also passed to subsequent networks, used for further accuracy enhancement guided by the text features.

\begin{figure*}[!t]
  \centering
  \includegraphics[width=0.9\textwidth]{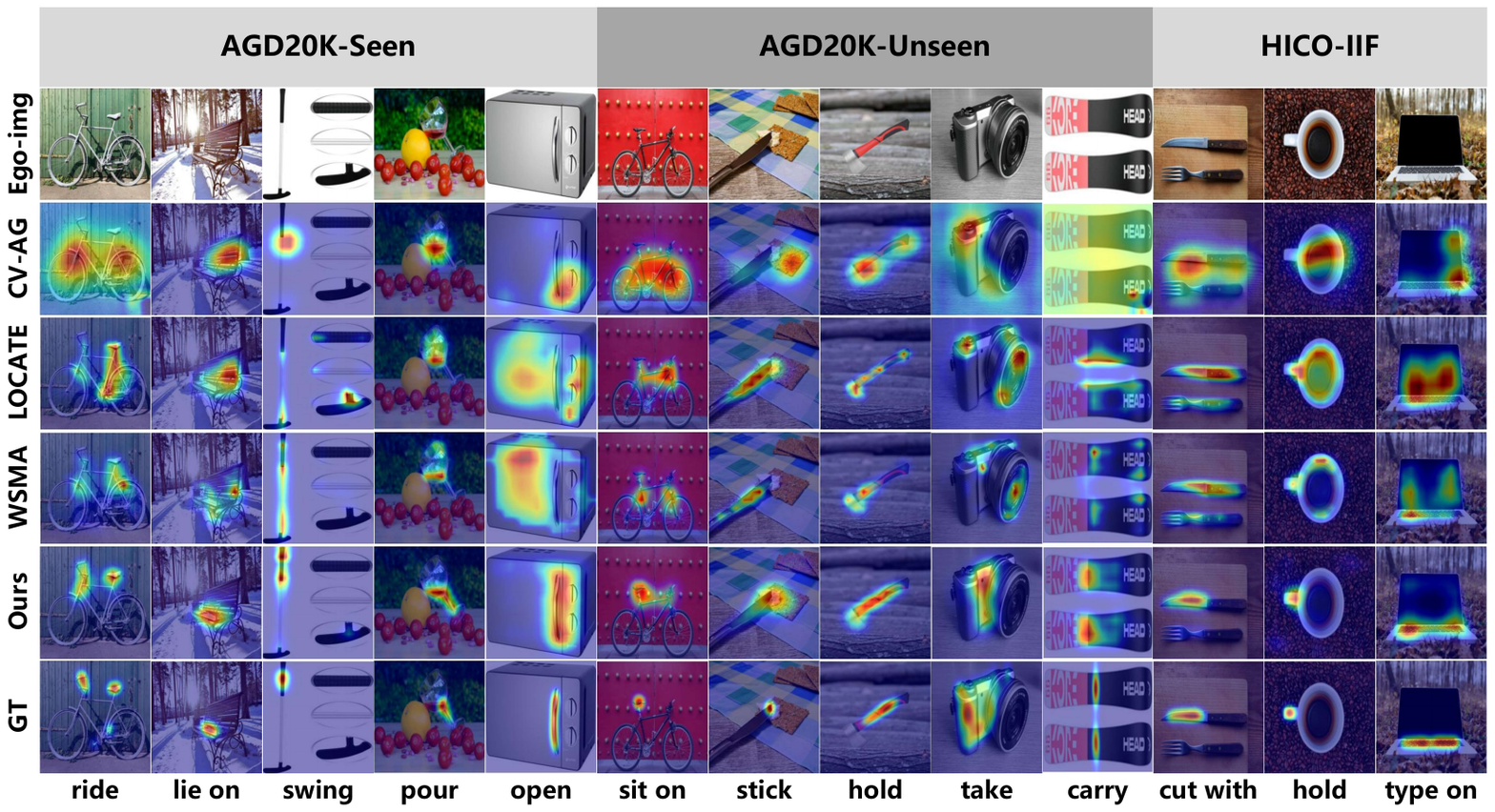}
  \vskip-14ex
  \caption{\textbf{Qualitative comparison of our proposed BiT-Align method with state-of-the-art affordance grounding approaches.} WSMA~\cite{xu2024weakly}, LOCATE~\cite{li2023locate}, and Cross-View-AG (CV-AG)~\cite{luo2022learning} are compared.   
  }
  \label{fig:qualitative}
  \vskip-3ex
\end{figure*}

\subsection{Text Feature Guidance}
\label{sec:text_feature_guidance}
 
To extract guidance information from the textual features, we utilize the $\Phi$ text encoder to obtain text embeddings $\mathcal{T}\in\mathbb{R}^{l \times d}$ of corresponding action labels. Subsequently, $\mathcal{N}$ learnable prompt tokens $\mathbf{p}\in\mathbb{R}^{d}$ are added, and the features are aligned through a linear projection:
\begin{equation}
\mathcal{F}_{text} = {Linear}({\Phi}({concat}(\mathbf{p},\mathcal{T}))).
\end{equation}

The image features from the egocentric branch pass through the text-guided self-attention module, which computes a similarity matrix between the CLS token and the text features $\mathcal{F}_{text}$. The matrix is also used to compute the cross-entropy loss $\mathcal{L}_{tcls}$ for optimization, helping the self-attention module better align the image and text features. Following the approach of PT module, the text feature vectors are expanded to the corresponding dimensions and undergo masked weighting with the image features row by row to further enhance the image-text association:
\begin{equation}
\mathcal{F}_{text}' = {Sigmoid}(\mu\mathcal{F}_{ego}  \mathcal{F}_{text}),
\end{equation}
\begin{equation}
\mathcal{F}_{p} = \mathcal{F}_{ego} \mathcal{F}_{text}' + \mathcal{F}_{ego}.
\end{equation}
Here, $\mu$ is used as a scaling factor to balance the difference between text and image. In the subsequent steps, we follow the approach outlined in~\cite{li2023locate}. Building on this, the attention weights obtained from the transformer encoder and self-attention modules can further guide the localization of affordance features. The text features obtained from the text encoder are processed through a simple MLP for a nonlinear mapping, where features are mapped to multiple categories corresponding to the attention heads CLS tokens $\mathcal{H}_{head}^{i}$ ($i{=}0,1,2,...,n$). Subsequently, the MLP outputs a sequence of weights ${h{\in}\mathbb{R}^{c \times n}}$ corresponding to each attention head. The attention heads serve as masks for the output features, and these are multiplied by the original output features to extract attention features $\mathcal{M}^{i}$ from the regions of interest for each attention head. Finally, the resulting attention features are multiplied by the weights output by the MLP, yielding a text-guided attention-weighted feature $\mathcal{F}_{t}$. The obtained features are added to the fused features $\mathcal{F}_{p}$ of the PT module. Among them, the coefficient $\alpha$ is used to balance the weights of the two fusion branches.
\begin{equation}
\mathcal{M}^{i} = \mathcal{H}_{head}^{i} \cdot \mathcal{F}_{ego},
\end{equation}
\begin{equation}
{h} = {MLP}\left(\mathcal{F}_{text} \right),
\end{equation}
\begin{equation}
\mathcal{F}_{t} = \frac{1}{n}\sum_{i=0}^{n}{h}^{i} \cdot \mathcal{M}^{i} ,
\end{equation}
\begin{equation}
\mathcal{F}_{f} = \alpha\mathcal{F}_{t} + (1-\alpha)\mathcal{F}_{p} .
\end{equation}

\subsection{Training and Inference}
During the training process, we use a cosine embedding loss $\mathcal{L}_{cos}$ to enforce the proximity between features from the exocentric and egocentric domains, and compensate for the domain gap using alpha. 
Following the approach of LOCATE~\cite{li2023locate} and the method in~\cite{hung2019scops}, we employ the concentration loss $\mathcal{L}_{c}$ to assist in generating more continuous prediction regions:
\begin{equation}
\mathcal{L} = \mathcal{L}_{cls} + \lambda_{tcls}\mathcal{L}_{tcls} + \lambda_{cos}\mathcal{L}_{cos} + \lambda_{c}\mathcal{L}_{c}.
\end{equation}
Here, $\lambda_{tcls}, \lambda_{cos}, and \lambda_{c}$ are the weighted coefficients for the loss, used to balance different components. $\mathcal{L}_{cls}$ represents the cross-entropy loss of the category features from the two branches after global average pooling.

In the testing phase, only the egocentric branch is retained, and RGB, depth, and textual label information are inputted to produce the activation map as the output. Since the exocentric branch encodes explicit knowledge, the model should learn to reason about possible actions solely from the egocentric branch, which lacks human actions. Due to robotic manipulator constraints (\textit{e.g.}, end-effector shape/size), tool interaction typically occurs at specific sub-regions rather than the entire operable area. Thus, we represent interaction targets via activation maps (not segmentation masks), concentrating predictions into focused peak regions with maximum probability density to guide precise end-effector positioning.

\begin{table*}[!t]
    \centering
    \resizebox{0.90\textwidth}{!}{
        \begin{tabular}{lc||ccc|ccc|ccc} 
            \hline\thickhline
             \rowcolor{mygray}
               & &\multicolumn{3}{c|}{AGD20K-Seen} &\multicolumn{3}{c|}{AGD20K-Unseen} &\multicolumn{3}{c}{HICO-IIF}\\
             \rowcolor{mygray}
             \multirow{-2}*{Method}& \multirow{-2}*{Pub.} &KLD\raisebox{0.2ex}{$\downarrow$} &SIM\raisebox{0.2ex}{$\uparrow$} &NSS\raisebox{0.2ex}{$\uparrow$} &KLD\raisebox{0.2ex}{$\downarrow$} &SIM\raisebox{0.2ex}{$\uparrow$} &NSS\raisebox{0.2ex}{$\uparrow$} &KLD\raisebox{0.2ex}{$\downarrow$} &SIM\raisebox{0.2ex}{$\uparrow$} &NSS\raisebox{0.2ex}{$\uparrow$}\\
            \hline\hline
            \textbf{Unimodal Methods} &&&&&&&&&\\
            {Cross-View-AG}~\cite{luo2022learning} & CVPR22  &1.538 &0.334 &0.927 &1.787 &0.285 &0.829 &1.779 &0.263 &0.946\\
            {LOCATE}~\cite{li2023locate}&  CVPR23  &1.226 &0.401 &1.177 &1.405 &0.372 &1.157 &1.593 &0.327 &0.966\\\hline
            
            \textbf{Multimodal Methods} &&&&&&&&&\\
            {World Affordance}~\cite{chen2024worldafford}& - &1.201 &0.406 &1.247 &1.393 &0.380 &1.225 &- &- &-\\
            {INTRA}~\cite{jang2025intra}&  ECCV24  &1.199 &0.407 &1.239 &1.365 &0.375 &1.209 &- &- &-\\
            {WSMA}~\cite{xu2024weakly}&   AAAI24  &1.176 &0.416 &1.247 &1.335 &\textbf{0.382} &1.220 &1.465 &0.358 &1.012\\
            {BiT-Align (Ours)}&           &\textbf{1.105} &\textbf{0.430} &\textbf{1.317} &\textbf{1.331}&0.371&\textbf{1.302}&\textbf{1.369} &\textbf{0.368} &\textbf{1.307}\\
            \hline
        \end{tabular}
    }
    \vspace{-1ex}
    \captionsetup{font=small}
    \caption{\small \textbf{Comparison of the proposed BiT-Align method with state-of-the-art approaches} on AGD20K~\cite{luo2022learning} and HICO-IIF~\cite{xu2024weakly}.}
    \label{table:comparison}
    \vspace{-4ex}
\end{table*}

\section{Experiments}

\subsection{Datasets and Data Processing}

We mainly use the public Affordance Grounding Dataset (AGD20K)~\cite{luo2022learning} and the HICO-IIF dataset~\cite{xu2024weakly} for evaluation.
AGD20K includes $20,061$ exocentric and $3,755$ egocentric images. 
The dataset consists of two subsets: `Seen' and `Unseen'. The former contains $36$ affordance labels, while the latter has $25$ affordance labels. The key difference is that the test set in `Unseen' includes categories not present in the training set. 
This dataset has been widely used to evaluate the performance of affordance methods. 
HICO-IIF includes $2,885$ exocentric and $1498$ egocentric images.
Yet, these datasets do not include depth data. 
In recent years, the emergence of monocular depth estimation methods with higher accuracy has provided a potential solution to compensate for missing depth data~\cite{yang2024depthanything,arampatzakis2023monocular,yang2024depthb}.
To assess the role of depth data and the effectiveness of fusion methods, we use Depth Anything V2~\cite{yang2024depthb} to generate corresponding pseudo-depth labels as depth image inputs.

In order to rationally measure the effectiveness of our methodology and compare it to previous methods, we use three metrics: Kullback-Leibler Divergence (KLD), Similarity (SIM), and Normalized Scanpath Saliency (NSS).

\subsection{Implementation Details}

For the image feature encoders of the egocentric and exocentric branches, we use the pre-trained DINO-ViT-S~\cite{oquab2023dinov2} and keep its weights frozen during training to control the overall model parameter count. 
DINO-ViT-S was pre-trained on the ImageNet dataset~\cite{deng2009imagenet} using self-supervised learning. In the exocentric branch, we adopt the same approach as LOCATE to simultaneously process inputs from three exocentric images. The $\lambda_{tcls}$, $ \lambda_{cos}$, and $\lambda_{c}$ are set to $(0.07, 1, 1)$.
The downscaling factor $\beta$ is set to $22$. 
For the text branch, we use the pre-trained text encoder of CLIP as the backbone network. 
Meanwhile, we employ the Pixel-Text fusion module from~\cite{xu2024weakly} to compute the contrastive loss between text and image features. The coefficient $\alpha$ is set to $0.8$. All experiments were conducted on an NVIDIA RTX 3090 GPU. The model was trained using Stochastic Gradient Descent (SGD) optimization with a batch size of $8$, initial learning rate of $1{\times}10^{-3}$, and weight decay coefficient of $5{\times}10^{-4}$. 
This configuration balances computational efficiency with stable convergence during both training and inference phases.

\subsection{Qualitative and Quantitative Results}
We compared our method with four previous approaches on the public AGD20K and HICO-IIF benchmarks, and visualized the results in Fig.~\ref{fig:qualitative}. 
Due to the incorporation of depth information, the proposed method demonstrates superior performance on transparent objects such as wine glasses in the fourth column. The integration of textual semantics enables the model to focus more on regions under the correct labels, such as bicycle handlebars in column one, while the TFG module helps suppress background noise. 
However, within the parts-level, this suppression is not as pronounced, as observed in the third column of the Unseen category,  where the TFG module fails to further suppress activations. Nevertheless, the predominantly activated regions are indeed relevant for the `hold' functionality.

\begin{table}[!t]
	\centering
	\small
	\resizebox{0.48\textwidth}{!}{
		\setlength\tabcolsep{8pt}
		\renewcommand\arraystretch{1.0}
		\begin{tabular}{l||c|c|c}
			\hline\thickhline
			\rowcolor{mygray}
			Methods  & \#Params (M)&GFLOPs & Time (ms)   \\ \hline\hline
            
            {Cross-View-AG}~\cite{luo2022learning}& 120 &29.18 &7.8  \\
			{LOCATE}~\cite{li2023locate} & 6.5 &5.0 & 5.5 \\\hline

			{World Affordance}~\cite{chen2024worldafford} &- &- &-  \\
			{INTRA}~\cite{jang2025intra} &- &-  &-   \\  
            {WSMA}~\cite{xu2024weakly} &68.6 &24.10 &12.8    \\ 
            {BiT-Align (Ours)} &7.7 & 6.64 & 10.0   \\ \hline
		\end{tabular}
	}
	\captionsetup{font=small}
    \vskip-1ex
	\caption{\small \textbf{Comparison of the number of trainable parameters, GFLOPs, and inference time} in various methods. LOCATE~\cite{li2023locate} is used as the baseline network.}
	\label{table:comparison02}
	\vspace{-3ex}
\end{table}

Comparisons with state-of-the-art methods~\cite{xu2024weakly,li2023locate,chen2024worldafford,jang2025intra}
demonstrate that our approach, while inheriting the low-parameter characteristic, integrates depth information and text prompts, achieving the best performance, as shown in Table~\ref{table:comparison}. 
Compared to unimodal methods, our multimodal approach only increases the number of trainable parameters by $19.2\%$, while achieving an $88.7\%$ reduction compared to RGB-D methods.
Additionally, we further investigated the advantages of the proposed method in terms of the inference speed. 
We conducted tests on an NVIDIA RTX3090 GPU.
After uniformly resizing the images to $224{\times}224$, we measured the GFLOPs of the model's inference branch. 
The inference time was calculated using CUDA calls, starting from the initiation and ending at the completion of the process. With a batch size of $1$, we recorded the inference time over $10$ runs and calculated the average value. 
The results are shown in Table~\ref{table:comparison02}. 
While the BPM module's block-wise processing of depth data increases inference time, its lightweight design limits this overhead to just $10.0$ ms faster than the public multimodal approach~\cite{xu2024weakly} which takes $12.8$ ms.

\begin{figure}[!t]
    \centering
    \includegraphics[width=0.48\textwidth]{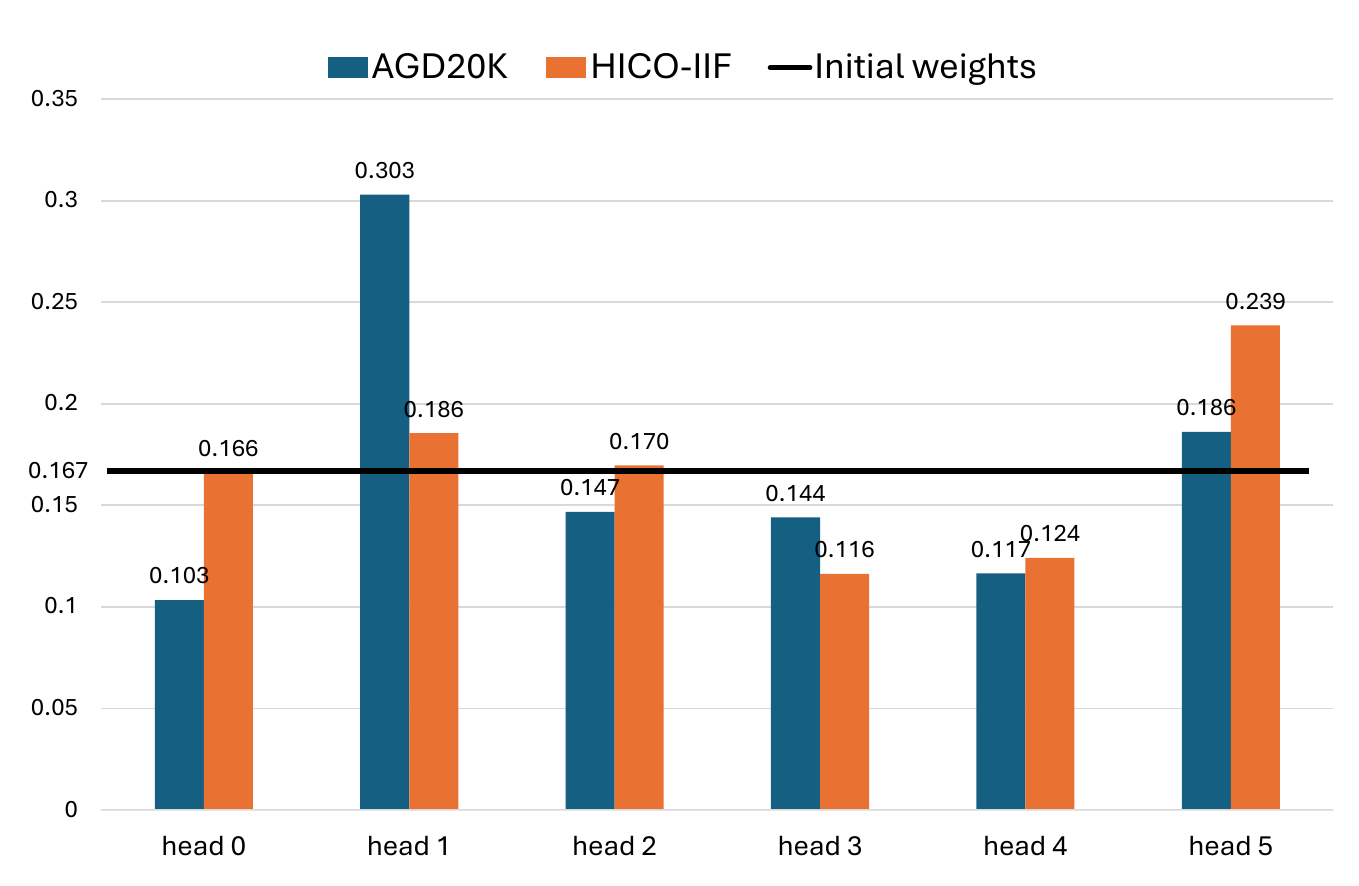} 
    \captionsetup{font=small}
    \vskip-2ex
    \caption{\textbf{Attention head weights after training.} The weight distributions are different across the two public datasets.}
    \label{fig:head_select}
    \vskip-2ex
\end{figure}

Moreover, we conducted a statistical analysis of the attention heads in DINO based on text input selection, and the results are shown in Fig.~\ref{fig:head_select}. It can be observed that the roles of different attention heads indeed exhibit certain distinctions depending on the text labels, which aligns with the findings described in~\cite{barsellotti2024talking}. 
Since DINO-ViT-S only uses $6$ attention heads and has a smaller scale, the discriminative ability at the parts level is limited.
However, the differences observed across the two datasets indicate that attention heads still help distinguish foreground from background in different environments.

\subsection{Ablation Studies}
In order to further explore the roles played by each part in the method we proposed, we first replaced the BPM module with the previous method in~\cite{jie2024convolutional, xu2024lv, dong2024efficient} to discuss the utility of the BPM module.
To verify the effectiveness of the fusion module in the bypass branch, we compared several other common prompt and fine-tuning methods in Table~\ref{table:comparison03}, demonstrating the advantages of our proposed module in handling multimodal fusion for affordance grounding.

\begin{table}[!t]
	\centering
	\small
	\resizebox{0.48\textwidth}{!}{
		\setlength\tabcolsep{8pt}
		\renewcommand\arraystretch{1.0}
		\begin{tabular}{l||c|ccc}
			\hline\thickhline
			\rowcolor{mygray}
			Methods & \#Params (M) & KLD{$\downarrow$} &SIM{$\uparrow$} &NSS{$\uparrow$}  \\ \hline\hline
			{Conv-adapter}~\cite{jie2024convolutional}&7.756  &1.901  &0.259  &0.429 \\%
		    {Lv-adapter}~\cite{xu2024lv}  &7.748  &1.400 &0.364 &1.049  \\  
                {MFP-adapter}~\cite{dong2024efficient} &7.748 & 1.138 &0.413 &1.307\\\hline
		\end{tabular}
	}
	\captionsetup{font=small}
    \vskip-1ex
	\caption{\small \textbf{Comparison of different branch fusion modules.} The BPM module is replaced with other prompt learning methods.}
	\label{table:comparison03}
	\vspace{-3ex}
\end{table}

Since BiT-Align has a side branch route composed of multiple inserted BPM modules, we further investigated the influence of the insertion position and quantity of the BPM modules, as well as whether the weights are shared or not, on the performance, as shown in Table~\ref{table:comparison04}. 
The number of BPM modules introduced and whether weights are shared will affect the number of trainable parameters in the model. 
When the number of BPM modules is small, the impact of weight sharing on both performance and parameter count is minimal. However, when the number of BPM modules is large, the model's performance is influenced by the structural placement of BPMs within the bypass prompt module. It can be confirmed that the best results are achieved when modules are used at each level with their weights being independent of each other. This also highlights that the affordance grounding task requires the utilization of multi-level feature representations, with varying demands across different levels. Ultimately, the fusion features composed of projections from multiple levels yield the best performance.

Finally, due to the simultaneous use of the PT module~\cite{xu2024weakly}, we conduct additional ablation experiments to evaluate its collaborative performance with the proposed TFG module.
The results are shown in Table~\ref{table:comparison05}. 
The PT module facilitates text localization but introduces background interference, whereas the TFG module counteracts this by dynamically enhancing foreground features through attention-head suppression of background noise, leading to improvements in KLD and SIM. 

\begin{figure}[!t]
  \centering
  \includegraphics[width=0.45\textwidth]{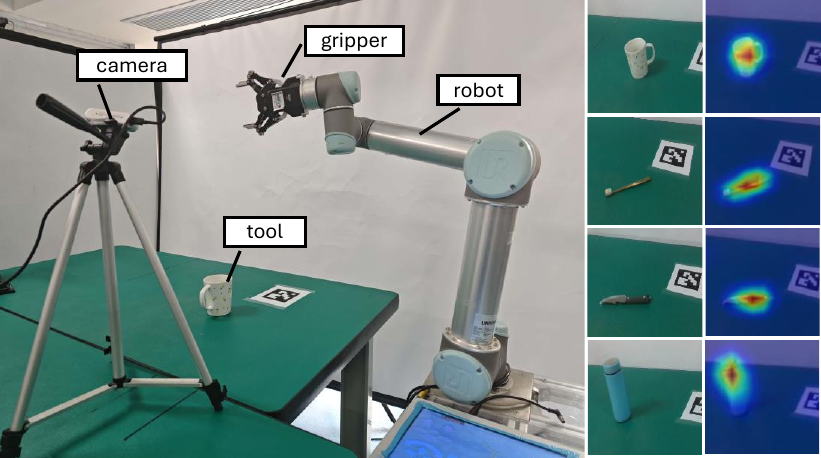}
  \vskip-1ex
  \caption{Experiment setup for real-world robotic applications. 
  }
  \label{fig:robot}
  \vskip -2ex
\end{figure}

\begin{table}[!t]
	\centering
	\small
	\resizebox{0.48\textwidth}{!}{
		\setlength\tabcolsep{8pt}
		\renewcommand\arraystretch{1.0}
		\begin{tabular}{l|c||c|ccc}
			\hline\thickhline
			\rowcolor{mygray}
			Blocks  & Share & \#Params (M)  & KLD{$\downarrow$} & SIM{$\uparrow$} & NSS{$\uparrow$} \\ \hline\hline
    
		  1,12   & \ding{51}  &7.465  &1.125 &0.427 &1.295 \\        
            1,5,9 & \ding{51}   &7.465  &1.124  &0.425  &1.296 \\
            1,4,7,10 & \ding{51}   &7.465 &1.126  &0.426 &1.296 \\
            1,3,5,7,9,11 & \ding{51}   &7.465 &1.130  &0.424 &1.303 \\
            1-12   & \ding{51}  &7.465  &1.127 &0.426 &1.298\\
            1,12   & \ding{55}  &7.490  &1.126 &0.425 &1.306 \\
            1,5,9 & \ding{55}   &7.516  &1.113  &0.428  &1.307 \\          
            1,4,7,10 & \ding{55}   &7.542 &1.137  &0.424  &1.287 \\
            1,3,5,7,9,11 & \ding{55}   &7.593 &1.110  &0.428  &1.316 \\
            1-12   & \ding{55}  &7.748  &1.105 &0.430 &1.317\\
            \hline
		\end{tabular}
	}
	\captionsetup{font=small}
    \vskip-1ex
	\caption{\small \textbf{Ablation study of the BPM module.} The blocks are placed in different positions for comparison.}%
	\label{table:comparison04}
	\vspace{-2ex}
\end{table}

\begin{table}[!t]
    \centering
    \small
        \begin{tabular}{ccc||ccc} 
            \hline\thickhline
             \rowcolor{mygray}
             {PT} & {TFG} &$\mathcal{L}_{tcls}$  & KLD{$\downarrow$} & SIM{$\uparrow$} & NSS{$\uparrow$} \\
            \hline\hline
             \ding{51}  &\ding{55}  &\ding{55}  &1.162 &0.410 &1.295\\
             \ding{55}  &\ding{51}  &\ding{55} &1.167 &0.420 &1.257  \\
             \ding{51}  &\ding{55}  &\ding{51}  &1.113 &0.420 &1.310\\
             \ding{55}  &\ding{51}  &\ding{51}  &1.130 &0.430 &1.301\\
            \hline
        \end{tabular}
    \captionsetup{font=small}
    \vskip-1ex
    \caption{\small \textbf{Ablation for the text fusion branch.} The text descriptions are derived from the class labels of the AGD20K dataset~\cite{luo2023learning}.}
    \label{table:comparison05}
    \vspace{-4ex}
\end{table}

\subsection{Insights Into Real-World Robotics Applications}

As shown in Fig.~\ref{fig:robot}, we validated our affordance grounding model on a real-world robotic manipulation platform comprising a UR5 robotic arm equipped with an Intel D435i RGB-D camera and an AG-95 gripper. 
The validation focused on verifying whether our parameter-efficient framework, which combines text labels with RGB-depth sensing from the Intel D435i, could achieve precise affordance localization for tool manipulation. Under controlled conditions with the UR5 arm's grasping posture and rotation angles constrained to a spatial plane, the model leveraged affordance predictions to identify interaction points for subsequent grasping decisions. 
Experimental results demonstrate that our method enabled 
successful grasping at functionally critical regions associated with the `hold' label (\textit{e.g.}, hammer handles, knife).
This validates that cross-modal grounding substantially enhances robotic operational precision.

\section{Conclusion}
In this work, we look into efficient multimodal affordance grounding and put forward the BiT-Align framework using image-depth-text affordance mapping.
The framework is constructed with the proposed Bypass Prompt Module (BPM) and Text Feature Guidance (TFG) mechanisms.
The proposed BiT-Align effectively integrates depth and semantic information and achieves state-of-the-art performance with only a small number of parameters, paving the way for more resource-efficient and accurate affordance learning and understanding in complex robotic environments.

In the future, we intend to further evaluate the framework's robustness to depth noise and domain shifts, while extending its application to more challenging real-world scenarios. Additionally, we will investigate deeper interactions between visual and textual encoders to enhance affordance reasoning.

{\small
\bibliographystyle{IEEEtran}
\bibliography{bib}

\begin{thebibliography}{10}
\providecommand{\url}[1]{#1}
\csname url@samestyle\endcsname
\providecommand{\newblock}{\relax}
\providecommand{\bibinfo}[2]{#2}
\providecommand{\BIBentrySTDinterwordspacing}{\spaceskip=0pt\relax}
\providecommand{\BIBentryALTinterwordstretchfactor}{4}
\providecommand{\BIBentryALTinterwordspacing}{\spaceskip=\fontdimen2\font plus
\BIBentryALTinterwordstretchfactor\fontdimen3\font minus \fontdimen4\font\relax}
\providecommand{\BIBforeignlanguage}[2]{{%
\expandafter\ifx\csname l@#1\endcsname\relax
\typeout{** WARNING: IEEEtran.bst: No hyphenation pattern has been}%
\typeout{** loaded for the language `#1'. Using the pattern for}%
\typeout{** the default language instead.}%
\else
\language=\csname l@#1\endcsname
\fi
#2}}
\providecommand{\BIBdecl}{\relax}
\BIBdecl

\bibitem{gibson1977theory}
J.~J. Gibson, \emph{The theory of affordances}.\hskip 1em plus 0.5em minus 0.4em\relax Hillsdale, NJ: Erlbaum, 1977, pp. 67--82.

\bibitem{puang2024learningstablerobotgrasping}
E.~Y. Puang, Z.~Li, C.~M. Chew, S.~Luo, and Y.~Wu, ``Learning stable robot grasping with transformer-based tactile control policies,'' \emph{arXiv preprint arXiv:2407.21172}, 2024.

\bibitem{delitzas2024scenefun3d}
A.~Delitzas, A.~Takmaz, F.~Tombari, R.~Sumner, M.~Pollefeys, and F.~Engelmann, ``{SceneFun3D:} {Fine-grained} functionality and affordance understanding in {3D} scenes,'' in \emph{Proc. CVPR}, 2024, pp. 14\,531--14\,542.

\bibitem{chuang2018learning}
C.~Chuang, J.~Li, A.~Torralba, and S.~Fidler, ``Learning to act properly: Predicting and explaining affordances from images,'' in \emph{Proc. CVPR}, 2018, pp. 975--983.

\bibitem{nagarajan2019grounded}
T.~Nagarajan, C.~Feichtenhofer, and K.~Grauman, ``Grounded human-object interaction hotspots from video,'' in \emph{Proc. ICCV}, 2019, pp. 8687--8696.

\bibitem{luo2023learning}
H.~Luo, W.~Zhai, J.~Zhang, Y.~Cao, and D.~Tao, ``Learning visual affordance grounding from demonstration videos,'' \emph{IEEE Transactions on Neural Networks and Learning Systems}, vol.~35, no.~11, pp. 16\,857--16\,871, 2024.

\bibitem{geng2023gapartnet}
H.~Geng \emph{et~al.}, ``{GAPartNet:} {Cross-category} domain-generalizable object perception and manipulation via generalizable and actionable parts,'' in \emph{Proc. CVPR}, 2023, pp. 7081--7091.

\bibitem{gao2024learning2d}
X.~Gao \emph{et~al.}, ``Learning {2D} invariant affordance knowledge for {3D} affordance grounding,'' \emph{Proc. AAAI}, 2025.

\bibitem{tong2024ovalprompt}
E.~Tong, A.~Opipari, S.~Lewis, Z.~Zeng, and O.~C. Jenkins, ``{OVAL-Prompt:} {Open-vocabulary} affordance localization for robot manipulation through {LLM} affordance-grounding,'' in \emph{Proc. ICRAW}, 2024.

\bibitem{xu2024weakly}
L.~Xu, Y.~Gao, W.~Song, and A.~Hao, ``Weakly supervised multimodal affordance grounding for egocentric images,'' in \emph{Proc. AAAI}, vol.~38, no.~6, 2024, pp. 6324--6332.

\bibitem{luo2022learning}
H.~Luo, W.~Zhai, J.~Zhang, Y.~Cao, and D.~Tao, ``Learning affordance grounding from exocentric images,'' in \emph{Proc. CVPR}, 2022, pp. 2242--2251.

\bibitem{li2023locate}
G.~Li, V.~Jampani, D.~Sun, and L.~Sevilla-Lara, ``{LOCATE:} {Localize} and transfer object parts for weakly supervised affordance grounding,'' in \emph{Proc. CVPR}, 2023, pp. 10\,922--10\,931.

\bibitem{koppula2013learning}
H.~S. Koppula, R.~Gupta, and A.~Saxena, ``Learning human activities and object affordances from {RGB-D} videos,'' \emph{The International Journal of Robotics Research}, vol.~32, no.~8, pp. 951--970, 2013.

\bibitem{kokic2017affordance}
M.~Kokic, J.~A. Stork, J.~A. Haustein, and D.~Kragic, ``Affordance detection for task-specific grasping using deep learning,'' in \emph{Proc. Humanoids}, 2017, pp. 91--98.

\bibitem{myers2015affordance}
A.~Myers, C.~L. Teo, C.~Ferm{\"{u}}ller, and Y.~Aloimonos, ``Affordance detection of tool parts from geometric features,'' in \emph{Proc. ICRA}, 2015, pp. 1374--1381.

\bibitem{hou2021affordance}
Z.~Hou, B.~Yu, Y.~Qiao, X.~Peng, and D.~Tao, ``Affordance transfer learning for human-object interaction detection,'' in \emph{Proc. CVPR}, 2021, pp. 495--504.

\bibitem{caron2021emerging}
M.~Caron \emph{et~al.}, ``Emerging properties in self-supervised vision transformers,'' in \emph{Proc. ICCV}, 2021, pp. 9630--9640.

\bibitem{luo2023leverage}
H.~Luo, W.~Zhai, J.~Zhang, Y.~Cao, and D.~Tao, ``Leverage interactive affinity for affordance learning,'' in \emph{Proc. CVPR}, 2023, pp. 6809--6819.

\bibitem{ragusa2023affordance}
E.~Ragusa, S.~Dosen, R.~Zunino, and P.~Gastaldo, ``Affordance segmentation using tiny networks for sensing systems in wearable robotic devices,'' \emph{IEEE Sensors Journal}, vol.~23, no.~19, pp. 23\,916--23\,926, 2023.

\bibitem{yang2023recent}
X.~Yang, Z.~Ji, J.~Wu, and Y.~Lai, ``Recent advances of deep robotic affordance learning: A reinforcement learning perspective,'' \emph{IEEE Transactions on Cognitive and Developmental Systems}, vol.~15, no.~3, pp. 1139--1149, 2023.

\bibitem{chu2019learning}
F.-J. Chu, R.~Xu, and P.~A. Vela, ``Learning affordance segmentation for real-world robotic manipulation via synthetic images,'' \emph{IEEE Robotics and Automation Letters}, vol.~4, no.~2, pp. 1140--1147, 2019.

\bibitem{chu2019toward}
F.~Chu, R.~Xu, L.~Seguin, and P.~A. Vela, ``Toward affordance detection and ranking on novel objects for real-world robotic manipulation,'' \emph{IEEE Robotics and Automation Letters}, vol.~4, no.~4, pp. 4070--4077, 2019.

\bibitem{chen2024worldafford}
C.~Chen, Y.~Cong, and Z.~Kan, ``{WorldAfford:} {Affordance} grounding based on natural language instructions,'' \emph{arXiv preprint arXiv:2405.12461}, 2024.

\bibitem{yoshida2024text}
T.~Yoshida, S.~Kurita, T.~Nishimura, and S.~Mori, ``Text-driven affordance learning from egocentric vision,'' \emph{arXiv preprint arXiv:2404.02523}, 2024.

\bibitem{jang2025intra}
J.~H. Jang, H.~Seo, and S.~Y. Chun, ``{INTRA:} {Interaction} relationship-aware weakly supervised affordance grounding,'' in \emph{Proc. ECCV}, vol. 15122, 2024, pp. 18--34.

\bibitem{li2024one}
G.~Li, D.~Sun, L.~Sevilla{-}Lara, and V.~Jampani, ``One-shot open affordance learning with foundation models,'' in \emph{Proc. CVPR}, 2024, pp. 3086--3096.

\bibitem{radford2021learning}
A.~Radford \emph{et~al.}, ``Learning transferable visual models from natural language supervision,'' in \emph{Proc. ICML}, vol. 139, 2021, pp. 8748--8763.

\bibitem{oquab2023dinov2}
M.~Oquab \emph{et~al.}, ``{DINOv2:} {Learning} robust visual features without supervision,'' \emph{Transactions on Machine Learning Research}, vol. 2024, 2024.

\bibitem{qian2024affordancellm}
S.~Qian, W.~Chen, M.~Bai, X.~Zhou, Z.~Tu, and L.~E. Li, ``{AffordanceLLM:} {Grounding} affordance from vision language models,'' in \emph{Proc. CVPR}, 2024, pp. 7587--7597.

\bibitem{li2024learning}
G.~Li \emph{et~al.}, ``Learning precise affordances from egocentric videos for robotic manipulation,'' \emph{arXiv preprint arXiv:2408.10123}, 2024.

\bibitem{yang2024depth}
L.~Yang, B.~Kang, Z.~Huang, X.~Xu, J.~Feng, and H.~Zhao, ``Depth anything: Unleashing the power of large-scale unlabeled data,'' in \emph{Proc. CVPR}, 2024, pp. 10\,371--10\,381.

\bibitem{xin2024parameter}
Y.~Xin \emph{et~al.}, ``Parameter-efficient fine-tuning for pre-trained vision models: A survey,'' \emph{arXiv preprint arXiv:2402.02242}, 2024.

\bibitem{jie2024convolutional}
S.~Jie, Z.~Deng, S.~Chen, and Z.~Jin, ``Convolutional bypasses are better vision transformer adapters,'' in \emph{Proc. ECAI}, vol. 392, 2024, pp. 202--209.

\bibitem{chen2022adaptformer}
S.~Chen \emph{et~al.}, ``{AdaptFormer:} {Adapting} vision transformers for scalable visual recognition,'' in \emph{Proc. NeurIPS}, vol.~35, 2022, pp. 16\,664--16\,678.

\bibitem{nie2023pro}
X.~Nie \emph{et~al.}, ``Pro-tuning: Unified prompt tuning for vision tasks,'' \emph{IEEE Transactions on Circuits and Systems for Video Technology}, vol.~34, no.~6, pp. 4653--4667, 2024.

\bibitem{xu2024lv}
G.~Xu, J.~Ye, X.~Liu, X.~Wen, Y.~Li, and J.~Wang, ``{Lv-Adapter:} {Adapting} vision transformers for visual classification with linear-layers and vectors,'' \emph{Computer Vision and Image Understanding}, vol. 246, p. 104049, 2024.

\bibitem{dong2024efficient}
S.~Dong, Y.~Feng, Q.~Yang, Y.~Huang, D.~Liu, and H.~Fan, ``Efficient multimodal semantic segmentation via dual-prompt learning,'' in \emph{Proc. IROS}, 2024, pp. 14\,196--14\,203.

\bibitem{hung2019scops}
W.~Hung, V.~Jampani, S.~Liu, P.~Molchanov, M.~Yang, and J.~Kautz, ``{SCOPS:} {Self-supervised} co-part segmentation,'' in \emph{Proc. CVPR}, 2019, pp. 869--878.

\bibitem{yang2024depthanything}
L.~Yang, B.~Kang, Z.~Huang, X.~Xu, J.~Feng, and H.~Zhao, ``Depth anything: Unleashing the power of large-scale unlabeled data,'' in \emph{Proc. CVPR}, 2024, pp. 10\,371--10\,381.

\bibitem{arampatzakis2023monocular}
V.~Arampatzakis, G.~Pavlidis, N.~Mitianoudis, and N.~Papamarkos, ``Monocular depth estimation: A thorough review,'' \emph{IEEE Transactions on Pattern Analysis and Machine Intelligence}, vol.~46, no.~4, pp. 2396--2414, 2024.

\bibitem{yang2024depthb}
L.~Yang \emph{et~al.}, ``Depth anything {V2},'' \emph{arXiv preprint arXiv:2406.09414}, 2024.

\bibitem{deng2009imagenet}
J.~Deng, W.~Dong, R.~Socher, L.-J. Li, K.~Li, and L.~Fei-Fei, ``{ImageNet:} {A} large-scale hierarchical image database,'' in \emph{Proc. CVPR}, 2009, pp. 248--255.

\bibitem{barsellotti2024talking}
L.~Barsellotti \emph{et~al.}, ``Talking to {DINO}: {Bridging} self-supervised vision backbones with language for open-vocabulary segmentation,'' \emph{arXiv preprint arXiv:2411.19331}, 2024.

\end{thebibliography}
}

\end{document}